\def\year{2019}\relax
\documentclass[letterpaper]{article} 
\usepackage{aaai19}  
\usepackage{times}  
\usepackage{helvet}  
\usepackage{courier}  
\usepackage{url}  
\usepackage{graphicx}  
\usepackage{xcolor}
\usepackage{float}
\usepackage{amsmath}
\usepackage{multirow}
\usepackage{kotex} 

\newcommand\Tstrut{\rule{0pt}{2.3ex}}         

\newcommand{\hh}[1]{\textcolor{black}{#1}}

\begin{document}
%
\title{Improving Neural Question Generation using Answer Separation}
\author{Yanghoon Kim\textsuperscript{1,2}, Hwanhee Lee\textsuperscript{1}, Joongbo Shin\textsuperscript{1} and Kyomin Jung\textsuperscript{1,2}\\
  \textsuperscript{1}Seoul National University, Seoul, Korea \\
  \textsuperscript{2}Automation and Systems Research Institute, Seoul National University, Seoul, Korea\\
  {\tt \{ad26kr,wanted1007,jbshin,kjung\}@snu.ac.kr} \\
}
\maketitle
\begin{abstract}
Neural question generation (NQG) is the task of generating a question from a given passage with deep neural networks. Previous NQG models suffer from a problem that a significant proportion of the generated questions include words in the question target, resulting in the generation of unintended questions. In this paper, we propose answer-separated seq2seq, which better utilizes the information from both the passage and the target answer. By replacing the target answer in the original passage with a special token, our model learns to identify which interrogative word should be used. 
We also propose a new module termed keyword-net, which helps the model better capture the key information in the target answer and generate an appropriate question. Experimental results demonstrate that our answer separation method significantly reduces the number of  improper questions which include answers. Consequently, our model significantly outperforms previous state-of-the-art NQG models.


\end{abstract}

\section{Introduction}

\noindent Neural question generation (NQG) is the task of generating questions from a given passage with deep neural networks. 
One of its key applications is to generate questions for educational materials \cite{heilman2010good}. It is also used as a way to improve question answering (QA) systems \cite{duan2017question,tang2017question,tang2018learning} or to engage chatbots to start and continue a conversation \cite{mostafazadeh2016generating}. 
%
%
%
%

\begin{figure}[t]
	\includegraphics[width = 0.48 \textwidth]{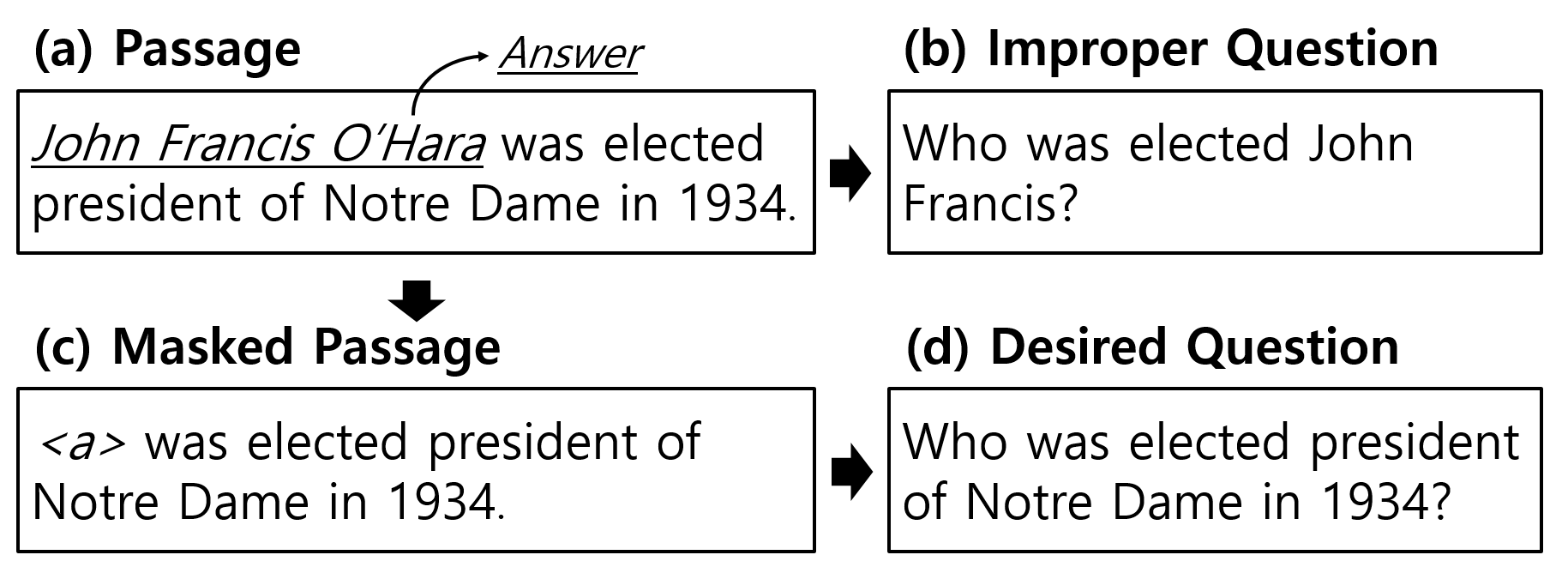}
  	\caption{
    An example of overall idea for QG in this paper. Generated questions from existing NQG models tend to include words from the answer, resulting in the generation of improper questions. Replacing the answer into a special token effectively prevents the answer words from appearing in the question, resulting in the generation of desired questions.
  }

\label{fig:replace_a_token}
\end{figure}

Automatic question generation (QG) from a passage is a challenging task due to the unstructured nature of textual data. One of major issues in NQG is how to take the question target, referred to as the target answer, in the passage. Specifying the question target is necessary for generating natural questions because there could be multiple target answers in the passage as in the following example. In Figure \ref{fig:replace_a_token}(a), the passage “John Francis O’Hara was elected president of Notre Dame in 1934.” has various candidates to be asked such as the person “John Francis O’Hara”, the location “Notre Dame”, and the number “1934.” Without taking the target answer as an additional input, existing NQG models such as \cite{du2017learning} tend to generate questions without specific target. This is a fundamental limitation due to the fact that recent NQG systems mostly rely on RNN sequence-to-sequence model \cite{sutskever2014sequence,bahdanau2014neural}, and RNNs do not have the ability to model high-level variability \cite{serban2017hierarchical}.
%
%
%
%


To overcome this limitation, most recent NQG models incorporated the target answer information by using the answer position feature \cite{zhou2017neural,song2018leveraging}. However, these approaches have a critical issue that a significant proportion of the generated questions include words in the target answer. For example, Figure \ref{fig:replace_a_token}(b) shows the improperly generated question “Who was elected John Francis?”\footnote{This example is actually generated by our base model which will be introduced in the later part.} which exposes some words in the answer. This problem results from the tendency of the sequence-to-sequence model to include all information from the passage \cite{amplayo2018entity}. It becomes severer with the recent trend that NQG models use the copy mechanism \cite{gulcehre2016pointing} to encourage that many words in the original passage appear in the question.

This study focuses on resolving this problem by separating the target answer from the original passage. For example, the masked passage  “\textless a\textgreater{} was elected president of Notre Dame in 1934.” in Figure \ref{fig:replace_a_token}(c) still contains enough information to generate the desired question in Figure \ref{fig:replace_a_token}(d), because the term “president” is mostly about someone's position. Interestingly, even though a target answer is replaced with a special token \textless a\textgreater{} in a passage, we can infer the interrogative word through contextual information from the remaining part of the passage. Therefore we expect that separating a target answer will prevent the answer inclusion problem.
%
%

In this paper, we develop a novel architecture named \textbf{answer-separated seq2seq} which treats the passage and the target answer separately for better utilization of the information from both sides. The first step in our NQG model is an answer masking. Literally, we replace the target answer with the mask token \textless a\textgreater{}, and keep the corresponding target answer apart. The masked passage is encoded by an RNN encoder inside of our model. This approach to separate the target answer from the passage helps our model to identify the question type related to the target answer because the model learns to capture the position and contextual information of the target answer with the help of the token \textless a\textgreater{}. Furthermore, we propose a new module called \textbf{keyword-net} as a part of answer-separated seq2seq, which extracts key information from the target answer kept apart before. The keyword-net makes our NQG model be consistently aware of the target answer, supplementing the information deficiency caused by answer separation. This module is inspired by how people keep the target answer in mind when they ask questions. Lastly, we adopt a retrieval-style word generator proposed by \cite{ma2018query} which better captures the word semantics during the generation process.
%
%

When we evaluate our answer-separated seq2seq on the SQuAD dataset \cite{rajpurkar2016squad}), our model outperforms previous state-or-the-art NQG models by a considerable margin. We empirically demonstrate the impact of the answer separation in following three ways: the rare appearance of the target answer in the generated questions, the better prediction of interrogative words, and the higher attention weights of the \textless a\textgreater{} token to interrogative words. Furthermore, trained with the only questions generated by our model, a machine comprehension system achieves a comparable results.
%
%
%
%


\section{Related Work}
Recently, there have been several NQG models which are end-to-end trainable from (\textit{passage}, \textit{question}, \textit{answer}) triplets written in natural language.
\cite{du2017learning} first dealt with end-to-end learning with regard to the question generation problem using a sequence-to-sequence model with an attention mechanism, achieving better performance than rule-based question generation methods in both automatic and human evaluations. 
However, their model did not take the target answer into account, resulting in generation of the questions which were full of randomness.

To generate more plausible questions, \cite{zhou2017neural} utilized answer positions to make the model aware of the target answer and used NER tags and POS tags as additional features. 
\cite{song2018leveraging} utilized the multi-perspective context matching algorithm of \cite{wang2017bilateral} to employ the interaction between the target answer and the passage for collecting the relevant contextual information. 
Both works employed a copy mechanism \cite{gulcehre2016pointing} to reflect the phenomenon by which many of the words in the original passage are copied to the generated question. 
However, none of them dealt with the issue of many of generated questions including target answers, and the copy mechanism could intensify this problem. 
To tackle this problem, this paper focuses on developing an NQG model that utilizes the target answer as a separated knowledge.

Additionally, there have been several works which utilize question generation to improve the question answering system.
\cite{duan2017question} crawled an external QA dataset and generated questions from it through their retrieval-based and generation-based question generation methods. With the generated questions as additional data for training the QA system, they demonstrated that their question generation model helps to improve QA systems.
More recently,\cite{tang2018learning} presented a joint training algorithm that improves both the question answering system and the question generation model.

To the best of our knowledge, none of the previous works has focused on the issue that a significant proportion of generated questions include words in the target answers.

\section{Task Definition}
Given a passage \(X^p = (x^p_1, ...,x^p_n)\) and a target answer \(X^a = (x^a_1, ..., x^a_m)\) as input, the NQG model aims to generate a question \(Y = (y_1, ...,y_T)\) asking about the target answer \(X^a\) in the passage \(X^p \). The NQG task is defined as finding the best \(\overline{Y}\) that maximizes the conditional likelihood given the \(X^p\) and the \(X^a\):

\begin{align}
\overline{Y} &= \underset{Y}{\operatorname{argmax}} P(Y|X^p, X^a) \label{eq_1} \\
& = \underset{Y}{\operatorname{argmax}} \sum_{t=1}^{T} P(y_t|X^p, X^a, y_{<t}) \label{eq_2}
\end{align}

\begin{figure*}[!htb]
	\centering
  	\includegraphics[width = \textwidth ]{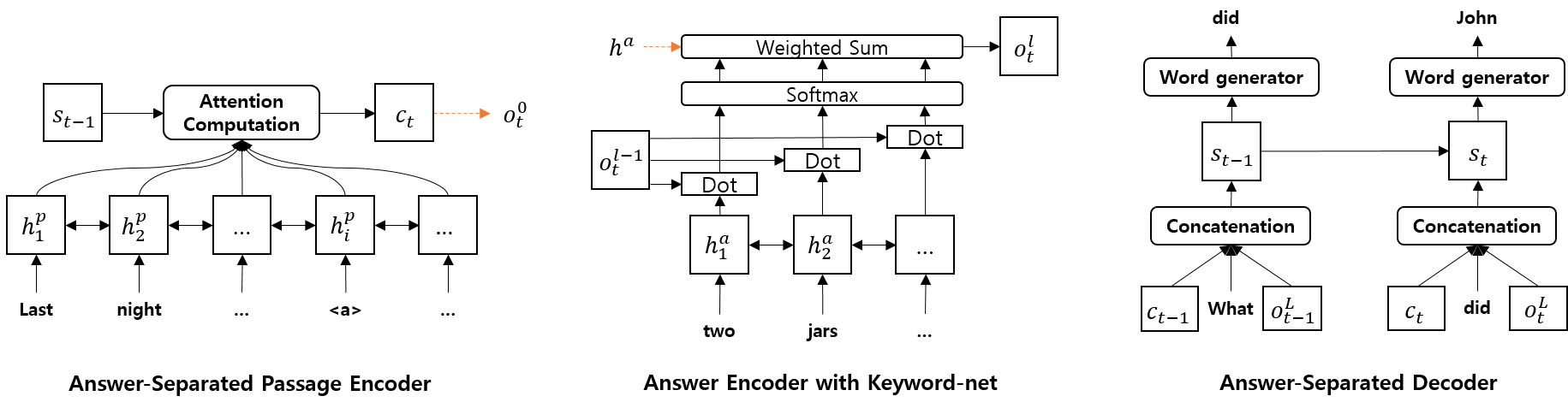}
	\caption{Overall architecture of our model}
  	\label{fig:overall_architecture}
\end{figure*}

\section{Base Model: Encoder-Decoder with Attention}
Following previous works, we base our model on the RNN encoder-decoder architecture \cite{sutskever2014sequence}, which is an RNN-based sequence-to-sequence learning model. It generates a task-specific sequential output from a given sequential input and is widely adopted in sequence generation tasks such as neural machine translation \cite{sutskever2014sequence,bahdanau2014neural}, text summarization \cite{nallapati2016abstractive} and dialogue model \cite{serban2016building,serban2017hierarchical}. In neural question generation, the model takes a passage \(X^p \) as an input and outputs a question \(Y\) which is relevant to the input passage \(X^p\). Note that the base model does not take the target answer as the input.

An RNN encoder-decoder model consists of two parts: an encoder and a decoder. The encoder is used to represent the variable-length input sequence as a fixed-length vector which includes contextual features of the input sequence, reflecting dependency among each input token. The decoder then generates an output sequence based on the encoder output.
%
%

In general, attention mechanism \cite{bahdanau2014neural}, which functions as visual attention mechanisms found in human, is combined with the encoder-decoder model. The mechanism alleviates the bottleneck that the decoder only relies on a fixed-size vector to generate sequences. With the attention mechanism, the decoder is able to pay attention to the most relevant parts of the given input sequence while generating an output sequence. In the following section, we describe the structure of the encoder-decoder with attention in details.
%
%
%

\subsubsection{Encoder}
The encoder is used to extract contextual features from the given input passage \(X^p\). We use an one-layer bi-directional LSTM as the encoder. A bi-directional LSTM consists of a forward LSTM and a backward LSTM:
\begin{gather}
\overset{\rightarrow}{h^p_i} = \overrightarrow{LSTM}(x^p_i, \overset{\rightarrow}{h^p}_{i-1})  \label{eq_3} \\
\overset{\leftarrow}{h^p_i} = \overleftarrow{LSTM}(x^p_i, \overset{\leftarrow}{h^p}_{i+1})  \label{eq_4} \\
h^p_i = [\overset{\rightarrow}{h^p_i};\overset{\leftarrow}{h^p_i}] \label{eq_5}
\end{gather}

For each time step \(i\), forward hidden state \(\overset{\rightarrow}{h^p_i}\) and backward hidden state \(\overset{\leftarrow}{h^p_i}\) are concatenated to form a hidden state of bi-LSTM. 

\subsubsection{Decoder}
Given the extracted features from the encoder, the decoder generates the corresponding question \(Y\). We utilize a one-layer uni-directional LSTM with attention:

\begin{gather}
s_t = LSTM(y_{t-1}, s_{t-1}, c_t) \label{eq_6} \\
P(y_t |y_{\textless{t}}, X^p) = Softmax(W_os_t) \label{eq_7}
\end{gather}

For each time step \(t\), the output token of previous time step \(y_{t-1}\), the hidden state of previous time step \(s_{t-1}\) and the context vector of current time \(c_t\) are passed through the decoder LSTM to compute the decoder hidden state of current time step \(s_t\). Each hidden state of decoder \(s_{t}\) is then linearly projected with a trainable weight matrix \(W_o\) and passed through a softmax layer to compute the probability of output \(y_t\). The context vector \(c_t\) is used to reflect the most relevant feature from the input passage \(X^p\) while generating the current question token \(y_t\).  In Eq. \eqref{eq_8},  the alignment score \(e_{ti}\) is computed as the matching score between \(s_{t-1}\) and \(h^p_i\), where \(W_c\) and \(U_c\) are trainable weight matrices and \(v^{\top}\) is a trainable vector. As in Eq. \eqref{eq_9}, the alignment weight \(\alpha_{ti}\) is computed with normalization and we take the weighted average of \(h^p_i\) as context vector:

\begin{gather}
e_{ti} = v^{\top}\tanh(W_cs_{t-1} + U_ch^p_i) \label{eq_8} \\
\alpha_{ti} = \frac{exp(e_{ti})}{\sum^{n}_{k=1}exp(e_{tk})} \label{eq_9} \\
c_t = \sum^{n}_{i=1}\alpha_{ti}h^p_i \label{eq_10}
\end{gather}

\section{Answer-Separated Seq2seq}
Previous encoder-decoder based neural question generation models take the whole passage \(X^p\) as an input. However, RNN encoders tend to pass all of the information in the passage to the decoder, causing a serious issue: the generated question often includes the target answer \(X^a\). Therefore, we propose \textbf{answer-separated seq2seq} which treats the target answer and the passage separately for better utilization of the information from both sides. With a simple pre-processing of data, we separate the target answer from the input passage. Encoded with two individual encoders of answer-separated seq2seq, contextual feature of the passage and the target answer are passed to the decoder. We further propose \textbf{keyword-net} as another part of answer-separated seq2seq, which is used to extract the key information from target answer.  In every decoding step, the decoder utilizes both the contextual feature of the passage from the attention mechanism and the keyword feature of the target answer from the keyword-net to generate a question that is related to the target answer in the passage. Furthermore, we adopt a retrieval style word generator by \cite{ma2018query} as the output layer of the decoder to better capture the word semantics.

Different from the general RNN encoder-decoder, our answer-separated seq2seq consists of :
\begin{itemize}
\item Two encoders each to extract contextual feature from the passage \(X^p\) and the target answer \(X^a\).
\item Answer-separated decoder which combines both the information from the passage and the target answer.
\end{itemize}

In the following section, we give a detailed description on how answer-separated seq2seq works. An overview of our answer-separated seq2seq is shown in Figure \ref{fig:overall_architecture}.

\subsection{Encoder}
Answer-separated seq2seq contains two individual encoders each for encoding the passage \(X^p\) and the target answer \(X^a\). Similar to the base model, we use two one-layer bi-LSTMs as encoders.

\subsubsection{Answer-Separated Passage Encoder}
Rather than feeding the passage encoder with additional features to emphasize the answer position, we first extract the target answer inside the passage and simply replace the corresponding target answer with a special \textless a\textgreater{} token as in Figure \ref{fig:replace_a_token}(c). In this way, the model learns to capture the position and contextual information of the target answer, knowing which part of the passage should be focused by the generated question. As a result, the probability that the generated question includes the target answer is reduced. This has a direct effect of preventing generation of questions irrelevant to the given target answer. We use the same one-layer bi-LSTMs as in Eq. \eqref{eq_1} and Eq. \eqref{eq_2}.

\subsubsection{Answer encoder}
We use another one-layer bi-LSTM to encode the target answer \(X^a\). In the last time step of the answer encoder, the hidden state of each LSTM is concatenated to form the final hidden state \(h^a_{final}\) , which represents the overall feature of the answer \(X^a\):

\begin{gather}
\overset{\rightarrow}{h^a_j} = \overrightarrow{LSTM}(x^a_j, \overset{\rightarrow}{h^a}_{j-1})  \label{eq_11} \\
\overset{\leftarrow}{h^a_j} = \overleftarrow{LSTM}(x^a_j, \overset{\leftarrow}{h^a}_{j+1})  \label{eq_12} \\
s_0 = h^a_{final} = [\overset{\rightarrow}{h^a_m};\overset{\leftarrow}{h^a_m}] \label{eq_13}
\end{gather}

\subsection{Answer-Separated Decoder}
To exploit sufficient information from both the passage and the target answer, we design the answer-separated decoder. Based on LSTM, answer-separated decoder employs features of the passage and the target answer in the following three ways.

\subsubsection{Decoder Initialization} 
We initialize the decoder state with the final answer vector \(h^a_{final}\).

\subsubsection{Incorporating the Key Feature of the Answer} 
We extract the key information in the target answer to disambiguate the question target. For example, given a passage “Steve Jobs is the founder of Apple” and the target answer “founder of Apple”, we want to generate a question like “Who is Steve Jobs?”. Then “founder” in “founder of Apple” is a keyword which defines the representative characteristic of the whole answer. In every decoding step, we use an attention-based module, termed \textbf{keyword-net}, to extract the key information from the target answer. For each layer of the keyword-net, a normalized matching score between output vector of last layer  \(o_t^{l-1}\) and answer hidden states \(h^a_j\) is computed. We then take the weighted average over \(h^a_j\) as the extracted keyword feature \(o^l_t\) in current layer \(l\). We initialize \(o^0_t\) with context vector \(c_t\) of current decoding step. Following equations describe the mechanism of keyword-net:
%
%

\begin{gather}
o^0_t = c_t \label{eq_14} \\
p^l_{tj} = Softmax((o^{l-1}_t)^{\top}h^a_{j}) \label{eq_15} \\
o_t^l = \sum_{j}p^l_{tj}h^a_{j}  \label{eq_16} \\
s_t = LSTM(y_{t-1}, s_{t-1}, c_t, o_t^L) \label{eq_17} 
\end{gather}

%
%

\subsubsection{Retrieval Style Word Generator} 
Based on the current decoder structure, we further adopt an architecture which can generate words by querying distributed word representation, with the purpose of capturing the semantic information of the according words.

\cite{ma2018query} proposed a retrieval style word generation layer which can remedy a shortcoming of the sequence-to-sequence model that sequence-to-sequence model has tendency to memorize the sequence pattern rather than reflecting word meanings. They made use of word embeddings to tackle the problem. Their word generator produces words by querying the distributed word representations, hoping to capture the meaning of used words. We then borrow the idea behind this novel word generator to replace the existing output layer in our decoder.

The query \(q_t\) is computed as a combination of the decoder hidden state \(s_t\) and the context vector \(c_t\). By querying \(q_t\) to each of the word embedding \(e_k\), we can compute the relevance score between \(q_t\) and \(e_k\) where \(W_a\) is a trainable parameter matrix. Then the normalized value of score function denotes the generation probability of each word. Since the original output layer takes the most of model parameters, we can dramatically reduce the parameter size and the time of model convergence by using this word retrieval layer:

\begin{gather}
q_t = \tanh(W_q[s_t;c_t]) \label{eq_18} \\
score(q_t, e_k) = q^\top_iW_ae_k \label{eq_19} \\
p(y_t) = Softmax(score(q_t, e_k)) \label{eq_20}
\end{gather}

\begin{table*}[!htb]
\centering
\begin{tabular}{|l|ccc|c|}
\hline
\multirow{2}{*}{Model} & \multicolumn{3}{c|}{Split-1} & Split-2\Tstrut  \\ \cline{2-5} 
 & BLEU & METEOR & ROUGE-L & BLEU\Tstrut \\  \hline
\cite{du2017learning} & 12.28 & 16.62 & 39.75 & - \Tstrut  \\
\cite{song2018leveraging} & 13.98 & 18.77 & 42.72 & 13.91  \\
\cite{zhou2017neural} & - & - & - & 13.29  \\ \hline
ASs2s-ASdec & 12.30 $\pm$ 0.26 & 16.70 $\pm$ 0.22 & 40.32 $\pm$ 0.26 & 12.25 $\pm$ 0.24 \Tstrut \\
ASs2s-keyword & 13.95 $\pm$ 0.29 & 19.34 $\pm$ 0.24 & 40.60 $\pm$ 0.25& 13.86 $\pm$ 0.30 \Tstrut \\
ASs2s-\textless a\textgreater{} & 14.37 $\pm$ 0.28 & 18.95 $\pm$ 0.24 & 42.06 $\pm$ 0.27 & 14.05 $\pm$ 0.30\Tstrut  \\
\textbf{ASs2s} & \textbf{16.20 $\pm$ 0.32} & \textbf{19.92$\pm$0.20} & \textbf{43.96 $\pm$ 0.25} & \textbf{16.17 $\pm$ 0.35}\Tstrut  \\ 
\hline
\end{tabular}%
\caption{Evaluation of our model and previous NQG models with three metrics: BLEU-4, METEOR and ROUGE-L.}
\label{table:evaluation}
\end{table*}

\section{Experimental Settings}
In this section, we first introduce the dataset we conduct experiments on. Then we give a detailed description of hyperparameter settings of our model. Lastly, several evaluation methods mainly used to assess the quality of generated questions are introduced. 

\subsection{Dataset}
For fair comparison, we use the same dataset that is used by previous works \cite{du2017learning,zhou2017neural,song2018leveraging}: two processed versions of SQuAD\cite{rajpurkar2016squad} dataset. The original SQuAD dataset contains 23,215 paragraphs from 536 articles with over 100k questions and their answers, which are originally created by crowd-workers. Since the original dataset is divided into train/dev splits, \cite{du2017learning,zhou2017neural} re-divided them into train/dev/test splits, and extracted passages from the paragraph that contains the target answer, each of which we call data split-1 and data split-2 in the following lines.  For the data split-1, since \cite{du2017learning} does not include the target answers, \cite{song2018leveraging} extracted them from each passage to make (\textit{passage}, \textit{question}, \textit{answer}) triplets. As a result, data split-1 and data split-2 contains 70,484/10,570/11,877 triplets and 86,635/8,965/8,964 triplets respectively. We tokenize both data splits with Stanford CoreNLP \cite{manning2014stanford} and then lower-case them.

\subsection{Implementation Details}
We implement our models in Tensorflow 1.4 and train the model with a single GTX 1080 Ti. The hyperparameters of our proposed model are described as follows.

Our model consists of two one-layer encoders each for encoding passages and target answers, and a one-layer decoder to generate questions. The number of hidden units in both encoders and the decoder are 350. For both encoder and decoder, we use 34k most frequent words appeared in training corpus, replacing the rest with the \textless UNK\textgreater\ token. We use 300-dimensional pre-trained GloVe \cite{pennington2014glove} embeddings trained on 6 billion-token corpus for initialization and freeze it when training. Weight normalization is applied to the attention module and dropout with \(P_{drop}\) = 0.4 is applied for both RNNs and the attention module. The layer size of keyword-net is set as 4. 

\subsubsection{Training and Inference}
During training, we optimize the cross-entropy loss function with the gradient descent algorithm using Adam \cite{kingma2014adam} optimizer, with an initial learning rate of 0.001. The mini-batch size for each update is set as 128 and the model is trained for up to 17 epochs. 

When testing, we conduct beam search with beam width 10 and length penalty weight 2.1. Decoding stops when the generated token is  \textless EOS\textgreater{}. The Performances of all our models are reported as mean and standard derivation values (Mean $\pm$ Std).
%
%


\subsubsection{Named Entity Replacement}
To further improve the model performance, we pre-process the data with a very simple technique. Since most named entities do not appear often, by replacing those named entities with representative tokens, we can not only reduce unknown words but also capture the grammatical structure. We look for the named entity tags for tokens in the given passage and replace each of them with the corresponding tag. We make sure that the same entity is assigned the same tag. NER tags are extracted with named entity tagger in Stanford CoreNLP. For those passages that have different named entities with the same tag, we distinguish them with different subscripts such as \(Person_1, Person_2\). We store a matching table between named entities and tags, which is used to post-process the generated questions.

\subsection{Evaluation Methods}
Following \cite{zhou2017neural,song2018leveraging}, we compare the performance of NQG models with 3 evaluation metrics: BLEU-4, Meteor and Rouge\(_L\), which are standard evaluation metrics of machine translation and text summarization. We use the evaluation package published by \cite{chen2015microsoft}.
%
%

\subsubsection{BLEU-\(4\)}
BLEU-\(4\) measures the quality of the candidate by counting the matching 4-grams in the candidate to the 4-grams in the reference text.

\subsubsection{Meteor}
Meteor compares the candidate with the reference in terms of exact, stem, synonym, and paraphrase matches between words and phrases.
%
%

\subsubsection{Rouge\(_L\)}
Rouge\(_L\) assesses the candidate based on longest common subsequence shared by both the candidate and the reference text.

\begin{table}[!tb]
\centering
\begin{tabular}{|l|c|c|}
\hline
Model        & Complete & Partial\Tstrut  \\ \hline
seq2seq+AP & 0.8\%            & 17.3\%\Tstrut               \\ \hline
\cite{song2018leveraging}  & 2.9\%        & 24.0\%\Tstrut          \\ \hline
ASs2s    & \textbf{0.6\%}        & \textbf{9.5\%}\Tstrut           \\ \hline
\end{tabular}
\caption{Percentage of complete/partial inclusion of the target answer in generated questions.}
\label{table:grasp answer}
\end{table}

\section{Results}

\begin{table*}[!htb]
\centering
\begin{tabular}{|l|l|l|l|l|l|l|l|l|}
\hline
\multirow{2}{*}{Model} & \multicolumn{8}{c|}{Question type}\Tstrut  \\ \cline{2-9} 
 & \multicolumn{1}{c|}{what} & \multicolumn{1}{c|}{how} & \multicolumn{1}{c|}{when} & \multicolumn{1}{c|}{which} & \multicolumn{1}{c|}{where} & \multicolumn{1}{c|}{who} & \multicolumn{1}{c|}{why} & \multicolumn{1}{c|}{yes/no}\Tstrut  \\ \hline
seq2seq+AP & 77.3\% & 56.2\% & 19.4\% & 3.4\% & 12.1\% & 36.7\% & 23.7\% & 5.3\%\Tstrut  \\ \hline
ASs2s & 82.0\% & 74.1\% & 43.3\% & 6.1\% & 46.3\% & 67.8\% & 28.5\% & 6.2\%\Tstrut  \\ \hline
\end{tabular}
\caption{Recall of interrogative word prediction.}
\label{table:interrogative word}
\end{table*}

\subsection{Performance Comparison}
We compare our model with previous state-of-the-art NQG models. Since there exists two different data splits processed by \cite{du2017learning,zhou2017neural}, we conduct experiments on both data splits. To figure out the effect of each module, we also conduct ablation tests against some key modules: \textbf{ASs2s} denotes the complete answer-separated seq2seq model. \textbf{ASs2s-\textless{} a\textgreater{}} is the answer-separated seq2seq without replacing the target answer in the original passage. \textbf{ASs2s-keyword} is the answer-separated seq2seq without keyword-net. \textbf{ASs2s-ASdec} is the answer-separated seq2seq without the answer-separated decoder but with a general LSTM decoder.

As shown in Table \ref{table:evaluation}, ASs2s outperforms all of the previous NQG models on both data splits by a great margin, showing that separate utilization of target answer information plays an important role in generating the intended questions. With the help of answer-separated decoder, ASs2s-\textless a\textgreater{} still outperforms the previous NQG models except for ROUGE-L on data split-1. However, there is a considerable decrease in all metrics compared to the complete model. This results from the fact that answer separation prevents generated question from including the answer. Similarly, ASs2s-keyword has a big drop in performance and this verifies that the keyword-net has actual impact on improving the performance. ASs2s-ASdec has greater decrease in all metrics compared to the ASs2s. \hh{This is a very natural result because without the answer-separated decoder, the model has to generate questions by only relying on context around the target answer position without knowledge of the target answer.}
%
%
%
%
%
%
%
%
%
%
%

\subsection{Impact of Answer Separation}
Answer separation helps the model generate  the right question for the given target answer. Since the base model does not utilize the target answer information, we further define \textbf{seq2seq+AP(Answer Position)} as base model with answer position feature \cite{zhou2017neural} for comparison. We show the benefits of answer-separated seq2seq in three aspects.

\subsubsection{Answer Copying Frequency}
%
%
If a NQG model captures the question target well, the generated question will rarely include the target answer. We verify the assumption by computing the percentage of generated questions including target answers. \hh{Since \cite{du2017learning} ignores the target answer, we choose seq2seq+AP to represent \cite{du2017learning} with answer position feature. } Further, we choose the previous state-of-the-art \cite{song2018leveraging} for comparison because both \cite{zhou2017neural} and \cite{song2018leveraging} use the copy mechanism. 
%
%
%
%
%
%

As shown in Table \ref{table:grasp answer}, \hh{the percentage that the target answers are either completely or partially included in the generated questions is significantly lower in our model.} We also \hh{figure out an interesting observation}: even though \cite{song2018leveraging} is the previous state-of-the-art NQG model, it generates more irrelevant questions to the target answer when compared to seq2seq+AP. This observation indicates the negative effect of copy mechanism that the target answer inside the passage is unintentionally copied to the generated question.
%
%
%
%

\subsubsection{Interrogative Word Prediction}
\hh{To figure out} the effect of answer-separated seq2seq on question type prediction, we compare the recall of each interrogative word prediction between the generated questions of answer-separated seq2seq and seq2seq+AP. We group questions into 8 categories: “what”, “how”, “when”, “which”, “where”, “who”, “why” and “yes/no”. As shown in Table \ref{table:interrogative word}, answer-separated seq2seq has better recall score over seq2seq+AP in all categories. Especially, the recall of question types “how”, “when”, “where” and “who” improved in big magnitude. Both model's recall of question type “what” is very high because “what” takes up more than half of the whole training set (55.4\%). Both model's recall of type “which” is very low. This may result from the fact that some combinations like “which year” and “which person” may be generated as “where” and “who” respectively. For question types “why” and “yes/no” which only take up 1.5\% and 1.2\% of the training set respectively, both models did not perform well due to the small amount of data.
%
%
%
%
%
%

\subsubsection{Attention from \textless a\textgreater{} }
We verify the effect of \hh{replacing answer with \textless a\textgreater{} by comparing attention matrices.} Given the passage  “john francis o'hara was elected president of notre dame in 1934.” and the target answer  “john francis o'hara”, following Figure \ref{fig:compare_attention}(a) and Figure \ref{fig:compare_attention}(c) show the attention matrices produced by our answer-separated seq2seq and seq2seq+AP respectively. 
%
%

\begin{figure*}[!htb]
	\centering
  	\includegraphics[width = \textwidth ]{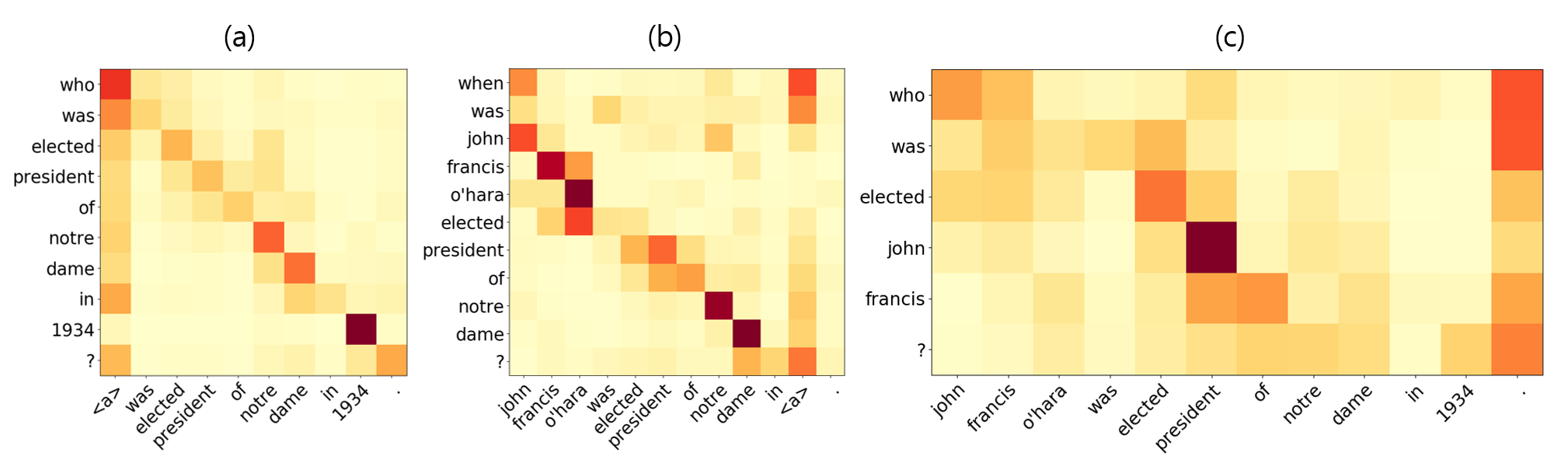}
	\caption{(a) and (b) show attention matrices of our model given a passage with two different target answers. (c) shows an attention matrix of seq2seq+AP given the same passage and the target answer as (a).}
  	\label{fig:compare_attention}
\end{figure*}

As shown in Figure \ref{fig:compare_attention}(a), the interrogative word "who" gets most of the attention weights(higher attention weights) from the \textless{}a\textgreater{} token in our answer-separated seq2seq. Further more, Our model can generate a question that is exactly related to the target answer. With additional answer position features as in Figure \ref{fig:compare_attention}(c), only a part of answer is attended while generating the interrogative word “who”. In this case, if the answer has some contextual information, then the model may omit it, generating an unintended question. Also, the generated question contains “john francis” which is a part of the target answer. We infer that the encoder tends to utilize more information from the word embeddings rather than answer position features, since the word embedding has far more information than answer position features.
%
%
%

\subsection{Question Generation for Machine Comprehension }
By training a machine comprehension system on the synthetic data generated by our model, we verify that our model has an enough ability to generate natural and fluent questions. By changing the position of the \textless a\textgreater{} token, we can easily produce various questions with our model. Figure \ref{fig:compare_attention}(a) and Figure \ref{fig:compare_attention}(b) shows one example where we use our model to generate two different questions corresponding to different target answers from the same input passage. 
%
%

We experiment with QANet \cite{yu2018qanet} on SQuAD dataset to verify whether the generated questions from our model are valid or not. Since most of the answers correspond to named entities, we use words and phrases that are named entities from training part of data split-1 as target answers. Then, we pair those answers with corresponding passages. We also make sure that selected answers are not overlapped with answers in the original SQuAD dataset because our NQG model is trained with the target answer provided with SQuAD dataset. If answers are overlapped, our model may generates exact the same questions as the golden questions. then we pair those answers with corresponding passages.
%
%
%

\hh{To organize the dataset in the same way as SQuAD dataset, (\textit{paragraph}, \textit{question}, \textit{answer position}) triplets, we trace the passage in data split-1 in the original paragraph and re-compute the answer position as well.} We finally make a synthetic data with about 50k questions and train the machine comprehension system only with our synthetic data.
%
%
%
%
%
As shown in Table \ref{table:qa}, the machine comprehension system achieves EM/F1 score of 22.72/31.58 in public SQuAD dev set. This result is far below the result 68.78/78.56 of the case when the model is trained with the original training set. However, considering our synthetic data only consists of target answers with single named entity, we further check EM/F1 score of partial dev set that only has a single named entity as the answer. We find that in the 10k dev set, about 40 percent of the data has an answer with a single named entity and the machine comprehension system achieves EM/F1 score of 49.09/56.57 with those parts of the data. Since the SQuAD dataset is a human-made dataset, this result sufficiently shows that our answer-separated seq2seq can generate valid questions that can be acceptable both by human and machine comprehension systems.
%
%
%
%


\begin{table}[!t]
\centering
\begin{tabular}{|l|c|c|}
\hline
Answers & Exact Match (EM)      & F1 score\Tstrut      \\ \hline
ALL     & 22.72       & 31.58\Tstrut       \\ \hline
NER     & 49.09 & 56.57\Tstrut \\ \hline
\end{tabular}%
\caption{Performance of the machine comprehension system which is trained only with synthetic data generated by our NQG model.}
\label{table:qa}
\end{table}

\section{Conclusion}
In this paper, we investigate the advantages of answer separation in neural question generation. We observe that existing NQG models suffer from a serious problem: a significant proportion of generated questions include words in the question target, resulting in the generation of unintended questions. To overcome this problem, we introduce a novel NQG architecture that treats the passage and the target answer separately to better utilize the information from the both sides. Experimental results show that our model has a strong ability to generate the right question for the target answer in the passage. As a result, it yields a substantial improvement over previous state-of-the-art models.

\section{Acknowledgments}
This work was supported by the National Research Foundation of Korea (NRF) funded by the Korean government (MSIT) (No. 2016M3C4A7952632), Industrial Strategic Technology Development Program (No. 10073144) funded by the Ministry of Trade, Industry \& Energy (MOTIE, Korea).


\bibliography{aaai}
\bibliographystyle{aaai}
\end{document}